\documentclass[sigconf, authorversion=true]{acmart}

\usepackage{booktabs} 
\usepackage{subcaption}

\setcopyright{rightsretained}

\copyrightyear{2017}
\acmYear{2017}
\setcopyright{acmcopyright}
\acmConference[GECCO '17]{the Genetic and Evolutionary Computation Conference 2017}{July 15-19, 2017}{Berlin, Germany}\acmPrice{15.00}\acmDOI{10.1145/3071178.3071229}
\acmISBN{978-1-4503-4920-8/17/07}

\begin{document}
\title[A Genetic Programming Approach to Designing CNN Architectures]{A Genetic Programming Approach to Designing Convolutional Neural Network Architectures}
\titlenote{The appendix and related references are added from the camera ready version.}

\author{Masanori Suganuma}
\affiliation{%
  \institution{Yokohama National University}
  \streetaddress{79-7 Tokiwadai Hodogaya-ku}
  \city{Yokohama} 
  \country{Japan} 
  \postcode{240-8501}
}
\email{suganuma-masanori-hf@ynu.jp}
\author{Shinichi Shirakawa}
\affiliation{%
  \institution{Yokohama National University}
  \streetaddress{79-7 Tokiwadai Hodogaya-ku}
  \city{Yokohama} 
  \country{Japan} 
  \postcode{240-8501}
}
\email{shirakawa-shinichi-bg@ynu.ac.jp}
\author{Tomoharu Nagao}
\affiliation{%
  \institution{Yokohama National University}
  \streetaddress{79-7 Tokiwadai Hodogaya-ku}
  \city{Yokohama} 
  \country{Japan} 
  \postcode{240-8501}
}
\email{nagao@ynu.ac.jp}

\begin{abstract}
The convolutional neural network (CNN), which is one of the deep learning models, has seen much success in a variety of computer vision tasks. However, designing CNN architectures still requires expert knowledge and a lot of trial and error. In this paper, we attempt to automatically construct CNN architectures for an image classification task based on Cartesian genetic programming (CGP). In our method, we adopt highly functional modules, such as convolutional blocks and tensor concatenation, as the node functions in CGP. The CNN structure and connectivity represented by the CGP encoding method are optimized to maximize the validation accuracy. To evaluate the proposed method, we constructed a CNN architecture for the image classification task with the CIFAR-10 dataset. The experimental result shows that the proposed method can be used to automatically find the competitive CNN architecture compared with state-of-the-art models.
\end{abstract}

%
%



\maketitle

\section{Introduction}

Deep learning, which uses deep neural networks as a model, has shown good performance on many challenging artificial intelligence and machine learning tasks, such as image recognition \cite{lecun_gradient-based_1998,krizhevsky_imagenet_2012}, speech recognition \cite{hinton_deep_2012}, and reinforcement learning tasks \cite{mnih_playing_2013,mnih_human-level_2015}.
In particular, convolutional neural networks (CNNs) \cite{lecun_gradient-based_1998} have seen huge success in image recognition tasks in the past few years and are applied to various computer vision applications \cite{vinyals_show_2015,zhang_colorful_2016}.
A commonly used CNN architecture consists mostly of several convolutions, pooling, and fully connected layers.
Several recent studies focus on developing a novel CNN architecture that achieves higher classification accuracy, e.g., GoogleNet \cite{szegedy_going_2015}, ResNet \cite{he_deep_2016}, and DensNet \cite{huang_densely_2016}.
Despite their success, designing CNN architectures is still a difficult task because many design parameters exist, such as the depth of a network, the type and parameters of each layer, and the connectivity of the layers.
State-of-the-art CNN architectures have become deep and complex, which suggests that a significant number of design parameters should be tuned to realize the best performance for a specific dataset.
Therefore, trial-and-error or expert knowledge is required when users construct suitable architectures for their target datasets.
In light of this situation, automatic design methods for CNN architectures are highly beneficial.

Neural network architecture design can be viewed as the model selection problem in machine learning. The straight-forward approach is to deal with architecture design as a hyperparameter optimization problem, optimizing hyperparameters, such as the number of layers and neurons, using black-box optimization techniques \cite{loshchilov_cma-es_2016,snoek_practical_2012}.

Evolutionary computation has been traditionally applied to designing neural network architectures \cite{schaffer_combinations_1992,stanley_evolving_2002}.
There are two types of encoding schemes for network representation: direct and indirect coding.
Direct coding represents the number and connectivity of neurons directly as the genotype, whereas indirect coding represents a generation rule for network architectures.
Although almost all traditional approaches optimize the number and connectivity of low-level neurons, modern neural network architectures for deep learning have many units and various types of units, e.g., convolution, pooling, and normalization.
Optimizing so many parameters in a reasonable amount of computational time may be difficult.
Therefore, the use of highly functional modules as a minimum unit is promising.

In this paper, we attempt to design CNN architectures based on genetic programming.
We use the Cartesian genetic programming (CGP) \cite{miller_cartesian_2000,harding_evolution_2008,miller_redundancy_2006} encoding scheme, one of the direct encoding schemes, to represent the CNN structure and connectivity.
The advantage of this representation is its flexibility; it can represent variable-length network structures and skip connections.
Moreover, we adopt relatively highly functional modules, such as convolutional blocks and tensor concatenation, as the node functions in CGP to reduce the search space.
To evaluate the architecture represented by the CGP, we train the network using a training dataset in an ordinary way. Then, the performance of another validation dataset is assigned as the fitness of the architecture. Based on this fitness evaluation, an evolutionary algorithm optimizes the CNN architectures.
To check the performance of the proposed method, we conducted an experiment involving constructing a CNN architecture for the image classification task with the CIFAR-10 dataset \cite{krizhevsky_learning_2009}. The experimental result shows that the proposed method can be used to automatically find the competitive CNN architecture compared with state-of-the-art models.

The rest of this paper is organized as follows. The next section presents related work on neural network architecture design. In Section 3, we describe our genetic programming approach to designing CNN architectures. We test the performance of the proposed approach through the experiment. Finally, in Section 5, we describe our conclusion and future work.

\section{Related Work}
This section briefly reviews the related work on automatic neural network architecture design: hyperparameter optimization, evolutionary neural networks, and the reinforcement learning approach.

\subsection{Hyperparameter Optimization}
We can consider neural network architecture design as the model selection or hyperparameter optimization problem from a machine learning perspective. There are many hyperparameter tuning methods for the machine learning algorithm, such as grid search, gradient search \cite{bengio_gradient-based_2000}, random search \cite{bergstra_random_2012}, and Bayesian optimization-based methods \cite{hutter_sequential_2011,snoek_practical_2012}. Naturally, evolutionary algorithms have also been applied to hyperparameter optimization problems \cite{loshchilov_cma-es_2016}. In the machine learning community, Bayesian optimization is often used and has shown good performance in several datasets. Bayesian optimization is a global optimization method of black-box and noisy objective functions, and it maintains a surrogate model learned by using previously evaluated solutions. A Gaussian process is usually adopted as the surrogate model \cite{snoek_practical_2012}, which can easily handle the uncertainty and noise of the objective function.
Bergstra et al. \cite{bergstra_algorithms_2011} have proposed the tree-structured Parzen estimator (TPE) and shown better results than manual search and random search.
They have also proposed a meta-modeling approach \cite{bergstra_making_2013} based on the TPE for supporting automatic hyperparameter optimization. 
Snoek et al. \cite{snoek_scalable_2015} used a deep neural network instead of the Gaussian process to reduce the computational cost for the surrogate model building and succeeded in improving the scalability.

The hyperparameter optimization approach often tunes predefined hyperparameters, such as the numbers of layers and neurons, and the type of activation functions. 
Although this method has seen success, it is hard to design more flexible architectures from scratch.

\subsection{Evolutionary Neural Networks}
Evolutionary algorithms have been used to optimize neural network architectures so far \cite{schaffer_combinations_1992,stanley_evolving_2002}. Traditional approaches are not suitable for designing deep neural network architectures because they usually optimize the number and connectivity of low-level neurons.

Recently, Fernando et al. \cite{fernando_convolution_2016} proposed differentiable pattern-producing networks (DPPNs) for optimizing the weights of a denoising autoencoder. The DPPN is a differentiable version of the compositional pattern-producing networks (CPPNs) \cite{stanley_compositional_2007}. This paper focuses on the effectiveness of indirect coding for weight optimization. That is, the general structure of the network should be predefined.

Verbancsics et al. \cite{verbancsics_generative_2013,verbancsics_image_2015} have optimized the weights of artificial neural networks and CNN by using the hypercube-based neuroevolution of augmenting topologies (HyperNEAT) \cite{stanley_hypercube-based_2009}. However, to the best of our knowledge, the methods with HyperNEAT have not achieved competitive performance compared with the state-of-the-art methods. Also, these methods require an architecture that has been predefined by human experts. Thus, it is hard to design neural network architectures from scratch.

\subsection{Reinforcement Learning Approach}
Interesting approaches, including the automatic designing of the deep neural network architecture using reinforcement learning, were attempted recently \cite{zoph_neural_2016,baker_designing_2016}.
These studies showed that the reinforcement learning-based methods constructed the competitive CNN architectures for image classification tasks.
In \cite{zoph_neural_2016}, a recurrent neural network (RNN) was used to generate neural network architectures, and the RNN was trained with reinforcement learning to maximize the expected accuracy on a learning task.
This method uses distributed training and asynchronous parameter updates with $800$ graphic processing units (GPUs) to accelerate the reinforcement learning process.
Baker et al. \cite{baker_designing_2016} have proposed a meta-modeling approach based on reinforcement learning to produce CNN architectures.
A Q-learning agent explores and exploits a space of model architectures with an $\epsilon -$greedy strategy and experience replay.

These approaches adopt the indirect coding scheme for the network representation, which optimizes generative rules for network architectures such as the RNN.
Unlike these approaches, our approach uses direct coding based on Cartesian genetic programming to design the CNN architectures.
In addition, we introduce relatively highly functional modules, such as convolutional blocks and tensor concatenations, to find better CNN architectures efficiently.

\begin{figure}[t]
\includegraphics[width=0.45\textwidth]{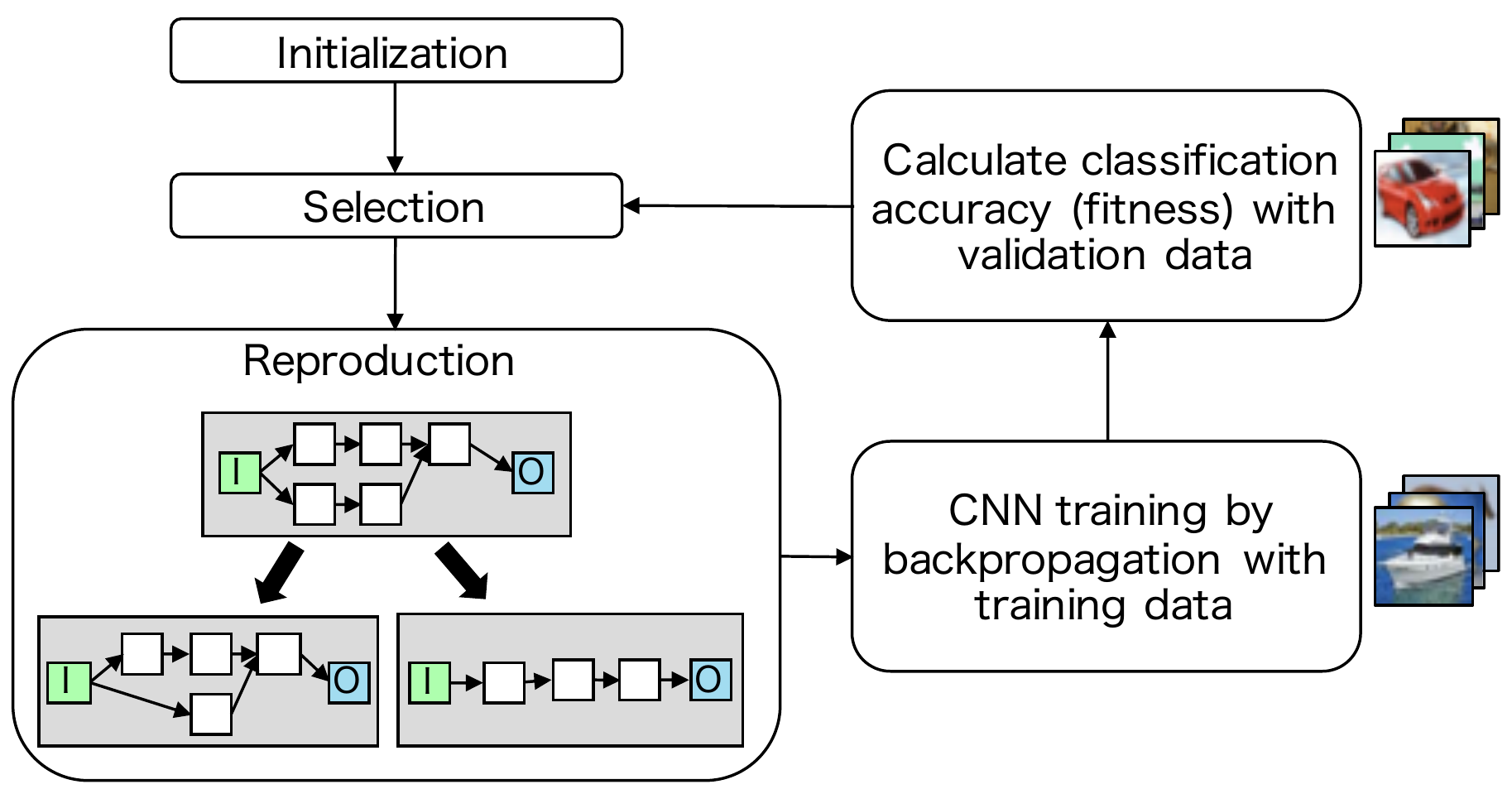}
\caption{Overview of our method. Our method represents CNN architectures based on Cartesian genetic programming. The CNN architecture is trained on a learning task and assigned the validation accuracy of the trained model as the fitness. The evolutionary algorithm searches the better architectures.}
\label{overview}
\end{figure}

\begin{figure*}[t]
\includegraphics[width=0.99\linewidth]{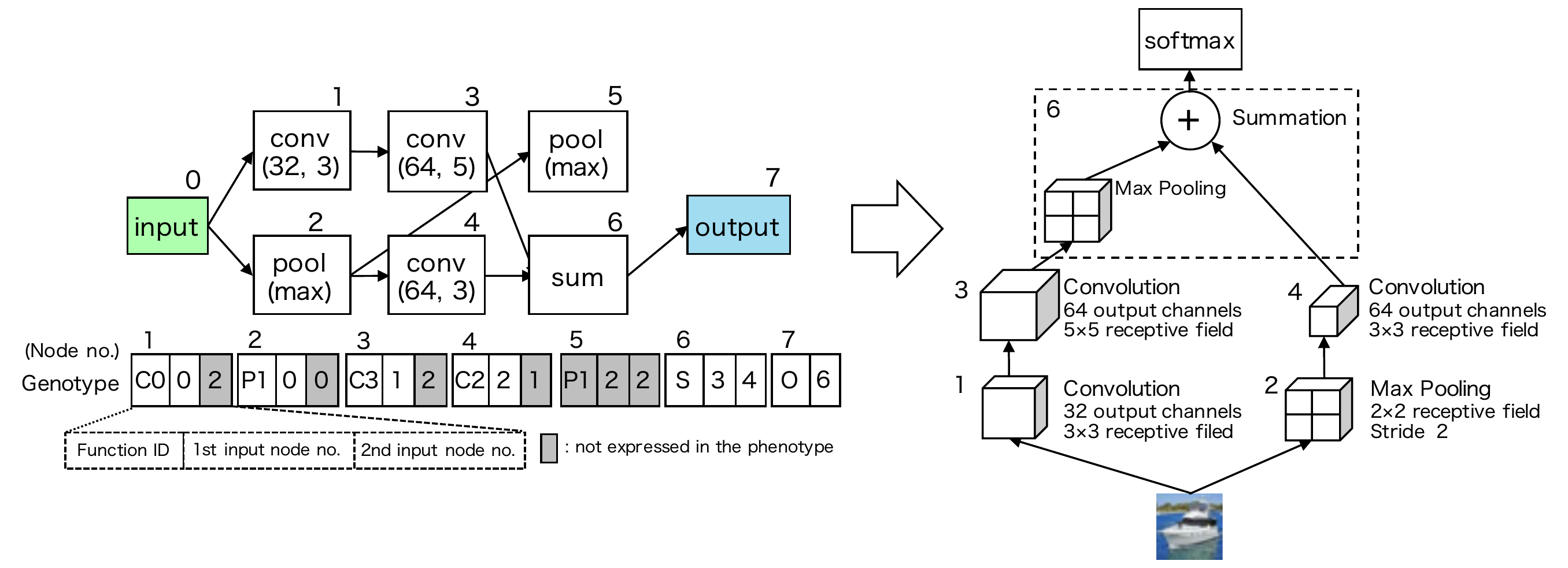}
\caption{Example of a genotype and a phenotype. The genotype (left) defines the CNN architecture (right). In this case, node No. 5 on the left side is an inactive node. The summation node performs max pooling to the first input so as to get the same input tensor sizes.}
\label{genotype}
\end{figure*}

\section{CNN Architecture Design Using Cartesian Genetic Programming}
Our method directly encodes the CNN architectures based on CGP \cite{harding_evolution_2008,miller_redundancy_2006,miller_cartesian_2000} and uses the highly functional modules as the node functions.
The CNN architecture defined by CGP is trained using a training dataset, and the validation accuracy is assigned as the fitness of the architecture. Then, the architecture is optimized to maximize the validation accuracy by the evolutionary algorithm.
Figure \ref{overview} illustrates an overview of our method.

In this section, we describe the network representation and the evolutionary algorithm used in the proposed method in detailed.

\subsection{Representation of CNN Architectures}
We use the CGP encoding scheme, representing the program as directed acyclic graphs with a two-dimensional grid defined on computational nodes, for the CNN architecture representation. Let us assume that the grid has $N_r$ rows by $N_c$ columns; then the number of intermediate nodes is $N_r \times N_c$, and the numbers of inputs and outputs depend on the task. The genotype consists of integers with fixed lengths, and each gene has information regarding the type and connections of the node. The $c$-th column's nodes should be connected from the $(c-l)$ to $(c-1)$-th column's nodes, where $l$ is called the levels-back parameter. Figure \ref{genotype} provides an example of the genotype, the corresponding network, and the CNN architecture in the case of two rows by three columns. Whereas the genotype in CGP is a fixed-length representation, the number of nodes in the phenotypic network varies because not all of the nodes are connected to the output nodes. Node No. 5 on the left side of Figure \ref{genotype} is an inactive node.

Referring to the modern CNN architectures, we select the highly functional modules as the node function.
The frequently used processings in the CNN are convolution and pooling; convolution processing uses a local connectivity and spatially shares the learnable weights, and pooling is nonlinear down-sampling.
We prepare the six types of node functions called ConvBlock, ResBlock, max pooling, average pooling, concatenation, and summation.
These nodes operate the three-dimensional (3-D) tensor defined by the dimensions of the row, column, and channel. Also, we call this 3-D tensor feature maps, where a feature map indicates a matrix of the row and column as an image plane.

The ConvBlock consists of standard convolution processing with a stride of 1 followed by batch normalization \cite{ioffe_batch_2015} and rectified linear units (ReLU) \cite{nair_rectified_2010}. In the ConvBlock, we pad the outside of input feature maps with zero values before the convolution operation so as to maintain the row and column sizes of the output. 
As a result, the $M \times N \times C$ input feature maps are transformed into $M \times N \times C'$ output ones, where $M$, $N$, $C$, and $C'$ are the numbers of rows, columns, input channels, and output channels, respectively.
We prepare several ConvBlocks with the different output channels and the receptive field size (kernel size) in the function set of CGP.

\begin{figure}[t]
\begin{center}
\includegraphics[scale=0.45]{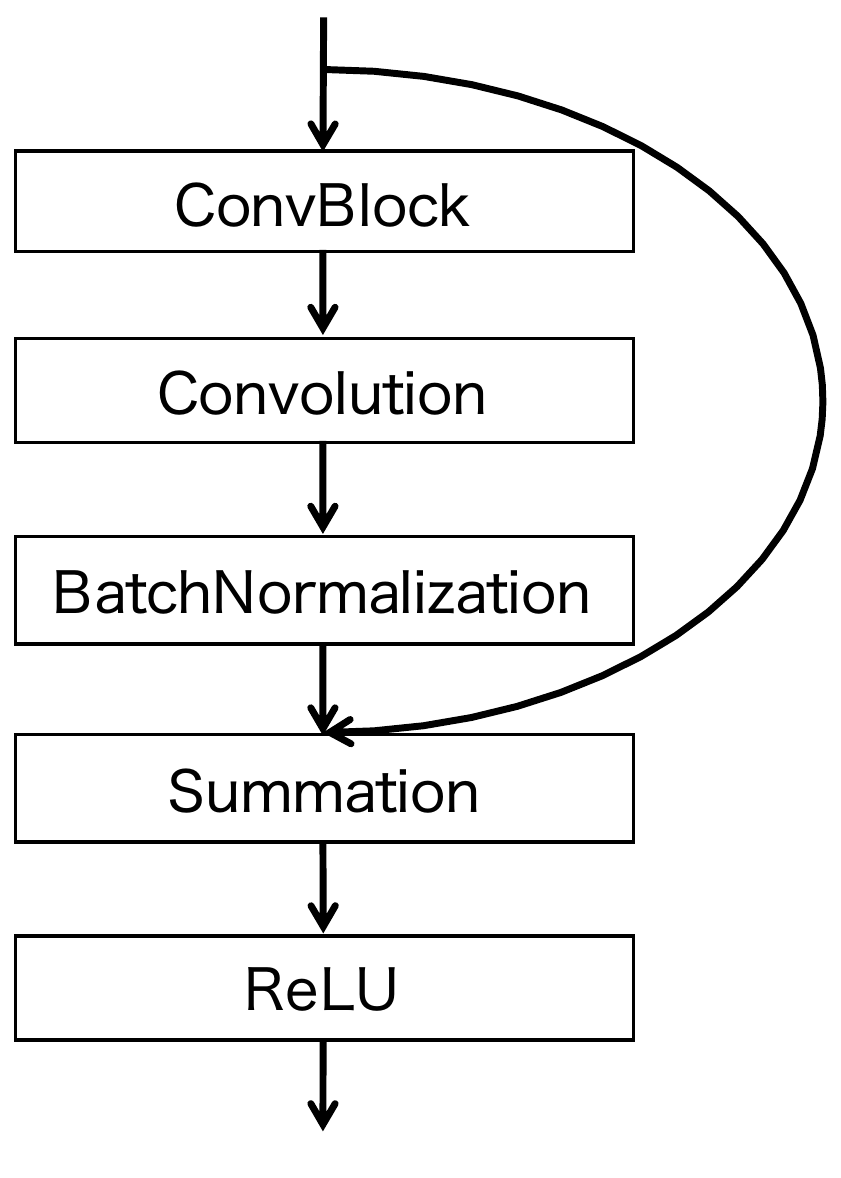}
\caption{ResBlock architecture.}
\label{resblock}
\end{center}
\end{figure}

The ResBlock is composed of a convolution processing, batch normalization, ReLU, and tensor summation. A ResBlock architecture is shown in Figure \ref{resblock}.
The ResBlock performs identity mapping by shortcut connections as described in \cite{he_deep_2016}.
The row and column sizes of the input are preserved in the same way as ConvBlock after convolution.
The output feature maps of the ResBlock are calculated by the ReLU activation and the summation with the input feature maps and the processed feature maps as shown in Figure \ref{resblock}.
In the ResBlock, the $M \times N \times C$ input feature maps are transformed into the $M\times N\times C'$ output ones.
We also prepare several ResBlocks with the different output channels and the receptive field size in the function set of CGP.

The max and average poolings perform a max and average operation, respectively, over the local neighbors of the feature maps. We use the pooling with the $2 \times 2$ receptive field size and the stride size 2.
In the pooling operation, the $M \times N \times C$ input feature maps are transformed into the $M' \times N' \times C$ output ones, where $M' = \lfloor M/2 \rfloor$ and $N' = \lfloor N/2 \rfloor$.

The concatenation function concatenates two feature maps in the channel dimension.
If the input feature maps to be concatenated have different numbers of rows or columns, we down-sample the larger feature maps by max pooling so that they become the same sizes of the inputs.
In the concatenation operation, the sizes of the output feature maps are $\min (M_1, M_2) \times \min (N_1, N_2) \times (C_1 + C_2)$, where as the sizes of the inputs are $M_1 \times N_1 \times C_1$ and $M_2 \times N_2 \times C_2$.

The summation performs the element-wise addition of two feature maps, channel by channel. 
In the same way as the concatenation, if the input feature maps to be added have different numbers of rows or columns, we down-sample the larger feature maps by max pooling.
In addition, if the inputs have different numbers of channels, we pad the smaller feature maps with zeros for increasing channels.
In the summation operation, the size of the output feature maps are $\min (M_1, M_2) \times \min (N_1, N_2) \times \max (C_1, C_2)$, where the sizes of the inputs are $M_1 \times N_1 \times C_1$ and $M_2 \times N_2 \times C_2$. In Figure \ref{genotype}, the summation node performs max pooling to the first input so as to get the same input tensor sizes. Adding these summation and concatenation operations allows our method to represent shortcut connections or branching layers, such as those used in GoogleNet \cite{szegedy_going_2015} and Residual Net \cite{he_deep_2016} without ResBlock.

The output node represents the softmax function with the number of classes. The outputs fully connect to all elements of the input.
The node functions used in the experiments are displayed in Table \ref{tbl:node_func}.

\subsection{Evolutionary Algorithm}
We use a point mutation as the genetic operator in the same way as the standard CGP. The type and connections of each node randomly change to valid values according to a mutation rate. 
The standard CGP mutation has the possibility of affecting only the inactive node. In that case, the phenotype (representing the CNN architecture) does not change by the mutation and does not require a fitness evaluation again.

The fitness evaluation of the CNN architectures is so expensive because it requires the training of CNN.
To efficiently use the computational resource, we want to evaluate some candidate solutions in parallel at each generation.
Therefore, we apply the mutation operator until at least one active node changes for reproducing the candidate solution. We call this mutation a forced mutation.
Moreover, to maintain a neutral drift, which is effective for CGP evolution \cite{miller_redundancy_2006,miller_cartesian_2000}, we modify a parent by the neutral mutation if the fitnesses of the offsprings do not improve.
In the neutral mutation, we change only the genes of the inactive nodes without the modification of the phenotype.

We use the modified $(1+\lambda)$ evolutionary strategy (with $\lambda = 2$ in our experiments) in the above artifice.
The procedure of our modified algorithm is as follows:
\begin{enumerate}
  \item Generate an initial individual at random as parent $P$, and train the CNN represented by $P$ followed by assigning the validation accuracy as the fitness.
  \item Generate a set of $\lambda$ offsprings $C$ by applying the forced mutation to $P$.
  \item Train the $\lambda$ CNNs represented by offsprings $C$ in parallel, and assign the validation accuracies as the fitness.
  \item Apply the neutral mutation to parent $P$.
  \item Select an elite individual from the set of $P$ and $C$, and then replace $P$ with the elite individual.
  \item Return to step $2$ until a stopping criterion is satisfied.
\end{enumerate}

\begin{table}[tb]
\caption{The node functions and abbreviated symbols used in the experiments.}
\label{tbl:node_func}
\begin{tabular}{l|l|l} \hline
Node type & Symbol & Variation \\ \hline
ConvBlock & CB ($C'$, $k$) & $C' \in \{32, 64, 128\}$ \\
                 &                      & $k \in \{ 3\times3, 5 \times 5 \}$ \\
ResBlock   & RB ($C'$, $k$) & $C' \in \{32, 64, 128\}$ \\
                 &                      & $k \in \{ 3 \times 3, 5 \times 5 \}$ \\
Max pooling       & MP              & -- \\
Average pooling & AP              & -- \\
Summation      & Sum     & -- \\
Concatenation & Concat & -- \\ \hline
\multicolumn{3}{l}{$C'$: Number of output channels} \\
\multicolumn{3}{l}{$k$: Receptive field size (kernel size)}
\end{tabular}
\end{table}

\section{Experiments and Results}
\subsection{Dataset}
We test our method on the image classification task using the CIFAR-10 dataset in which the number of classes is $10$. The numbers of training and test images are $50,000$ and $10,000$, respectively, and the size of images is $32 \times 32$.

We consider two experimental scenarios: the default scenario and the small-data scenario.
The default scenario uses the default numbers of the training images, whereas the small-data scenario assumes that we use only $5,000$ images as the learning data. 

In the default scenario, we randomly sample $45,000$ images from the training set to train the CNN, and we use the remaining $5,000$ images for the validation set of the CGP fitness evaluation.
In the small-data scenario, we randomly sample $4,500$ images for the training and $500$ images for the validation.

\begin{table}[t]
  \caption{Parameter setting for the CGP}
  \label{cgp_param}
  \begin{tabular}{l|l} \hline
    Parameters & Values \\ \hline
   Mutation rate & $0.05$ \\
   \#  Rows ($N_r$) & $5$ \\
   \#  Columns ($N_c$) & $30$ \\
   Levels-back ($l$) & $10$ \\ \hline
  \end{tabular}
\end{table}

\subsection{Experimental Setting}
To assign the fitness to the candidate CNN architectures, we train the CNN by stochastic gradient descent (SGD) with a mini-batch size of $128$. The softmax cross-entropy loss is used as the loss function.
We initialize the weights by the He's method \cite{he_delving_2015} and use the Adam optimizer \cite{kingma_adam:_2015} with $\beta_{1}=0.9$, $\beta_{2}=0.999$, $\varepsilon=1.0 \times 10^{-8}$, and an initial learning rate of $0.01$.
We train each CNN for 50 epochs and reduce the learning rate by a factor of 10 at 30th epoch.

We preprocess the data with the per-pixel mean subtraction.
To prevent overfitting, we use a weight decay with the coefficient $1.0 \times 10^{-4}$ and data augmentation.
We use the data augmentation method based on \cite{he_deep_2016}: padding $4$ pixels on each side followed by choosing a random $32\times 32$ crop from the padded image, and random horizontal flips on the cropped $32 \times 32$ image.

The parameter setting for CGP is shown in Table \ref{cgp_param}. We use the relatively larger number of columns than the number of rows to generate deep architectures that are likely.
The offspring size of $\lambda$ is set to two; that is the same number of GPUs in our experimental machines.
We test two node function sets called ConvSet and ResSet for our method.
The ConvSet contains ConvBlock,  Max pooling, Average pooling, Summation, and Concatenation in Table \ref{tbl:node_func}, and the ResSet contains ResBlock, Max pooling, Average pooling, Summation, and Concatenation.
The difference between these two function sets is whether we adopt ConvBlock or ResBlock.
The numbers of generations are $500$ for ConvSet, $300$ for ResSet in the default scenario, and $1,500$ in the small-data scenario, respectively.

After the CGP process, we re-train the best CNN architecture using each training image ($50,000$ for the default scenario and $5,000$ for the small-data scenario), and we calculate the classification accuracy for the $10,000$ test images to evaluate the constructed CNN architectures.

In this re-training phase, we optimize the weights of the obtained architecture for $500$ epochs with a different training procedure; we use SGD with a momentum of $0.9$, a mini-batch size of $128$, and a weight decay of $5.0 \times 10^{-4}$.
We start a learning rate of $0.01$ and set it to $0.1$ at $5$th epoch, then we reduce it to $0.01$ and $0.001$ at $250$th and $375$th epochs, respectively. This learning rate schedule is based on the reference \cite{he_deep_2016}.

We have implemented our methods using the Chainer \cite{tokui_chainer:_2015} (version 1.16.0) framework and run the experiments on the machines with $3.2$GHz CPU, $32$GB RAM, and two NVIDIA GeForce GTX 1080 (or two GTX 1070) GPUs.
It is possible that the large architectures generated by the CGP process cannot run in the environment due to the GPU memory limitation.
In that case, we assign a zero fitness to the candidate solution.

\begin{table}[t]
  \caption{Comparison of error rates on the CIFAR-10 dataset (default scenario). The values of Maxout, Network in Network, ResNet, MetaQNN, and Neural Architecture Search are referred from the reference papers.}
  \label{results}
  \begin{tabular}{l|c|c} \hline
   Model & Error rate & \# params ($\times 10^6$) \\ \hline
   Maxout \cite{goodfellow_maxout_2013} & $9.38$ & -- \\ 
   Network in Network \cite{lin_network_2014} & $8.81$ & -- \\
   VGG \cite{simonyan_very_2014} \footnotemark & $7.94$ & $15.2$ \\
   ResNet \cite{he_deep_2016} & $6.61$ & $1.7$ \\
   MetaQNN \cite{baker_designing_2016} \footnotemark & $9.09$ & $3.7$ \\
   Neural Architecture Search \cite{zoph_neural_2016} & $3.65$ & $37.4$ \\
   CGP-CNN (ConvSet) & $6.75$ & $1.52$ \\
   CGP-CNN (ResSet) & $5.98$ & $1.68$ \\ \hline
  \end{tabular}
\end{table}
\footnotetext[1]{We have implemented the VGG net \cite{simonyan_very_2014} for the CIFAR-10 dataset because the VGG net is not applied to the CIFAR-10 dataset in \cite{simonyan_very_2014}. The architecture of the VGG is identical with configuration D in \cite{simonyan_very_2014}. We denote this model as VGG in this paper.}
\footnotetext[2]{The mean error rate and the number of parameters of the top five models are shown.}

\begin{figure*}[t]
 \begin{minipage}[b]{0.45\linewidth}
  \centering
  \includegraphics[keepaspectratio, scale=0.4]{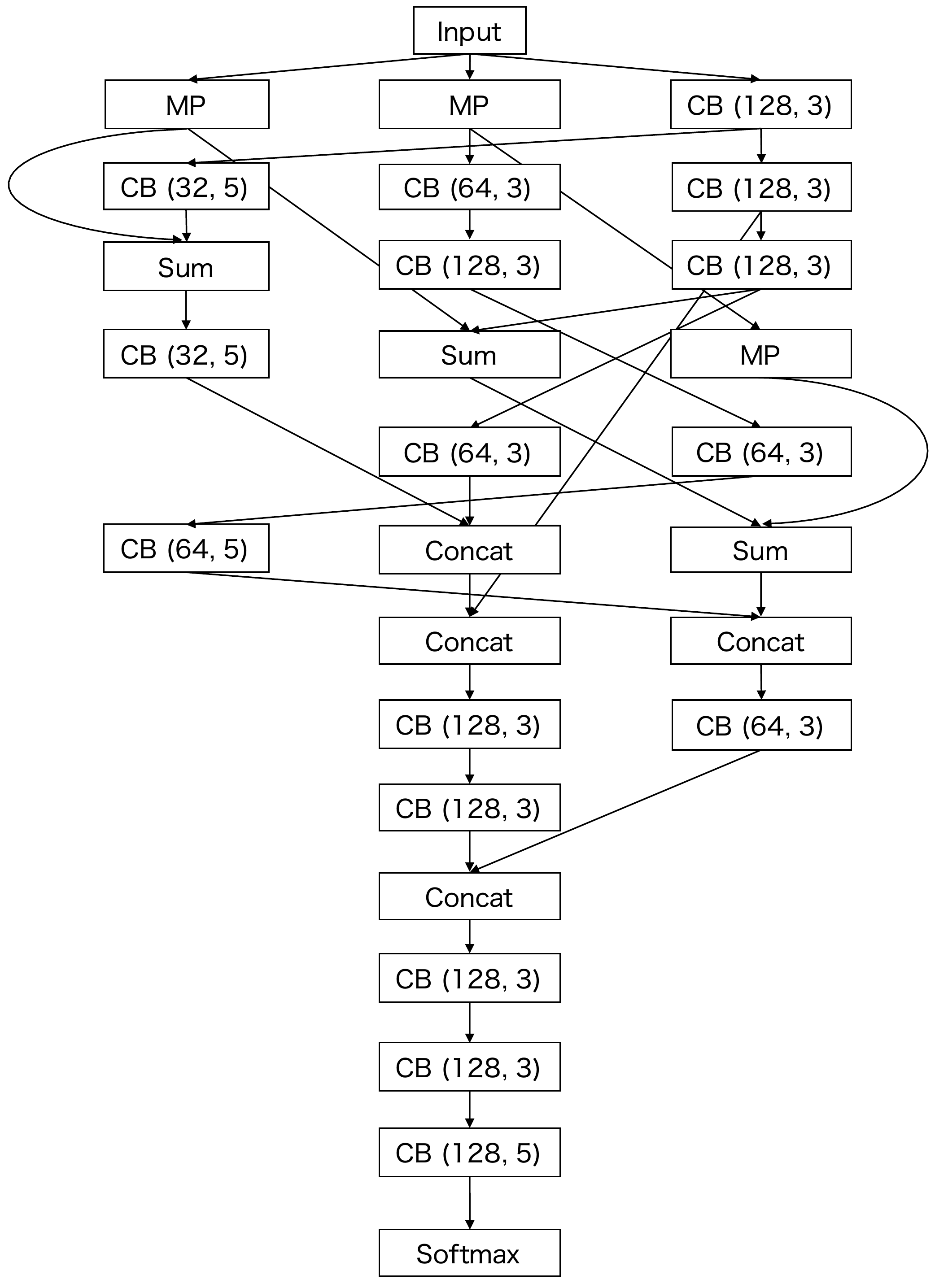}
  \subcaption{CGP-CNN (ConvSet)}\label{modelA}
 \end{minipage}
 \begin{minipage}[b]{0.45\linewidth}
  \centering
  \includegraphics[keepaspectratio, scale=0.4]{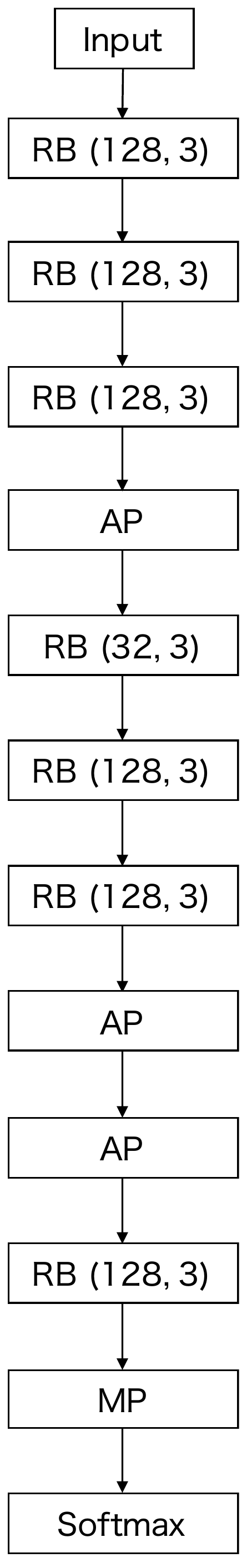}
  \subcaption{CGP-CNN (ResSet)}\label{modelB}
 \end{minipage}
 \caption{The CNN architectures designed by our method on the default scenario.}\label{models}
\end{figure*}

\subsection{Result of the Default Scenario}
We compare the classification performance of our method with the state-of-the-art methods and summarize the classification error rates in Table \ref{results}. We refer to the architectures constructed by the proposed method as CGP-CNN.
For instance, CGP-CNN (ConvSet) means the proposed method with the node function set of ConvSet.
The models, Maxout, Network in Network, VGG, and ResNet, are hand-crafted CNN architectures, whereas the MetaQNN and Neural Architecture Search are the models constructed by the reinforcement learning-based method.
Hand-crafted CNN architectures mean the CNN architectures are designed by human experts.
In Table \ref{results}, the numbers of learnable weight parameters in the models are also listed.

As can be seen in Table \ref{results}, the error rates of our methods are competitive with the state-of-the-art methods.
In particular, CGP-CNN (ResSet) outperforms all hand-crafted models, and the architectures constructed by using our method have a good balance between classification errors and the number of parameters. The Neural Architecture Search achieved the best error rate, but this method used 800 GPUs for the architecture search. Our method could find a competitive architecture with a reasonable machine resource.

Figure \ref{models} shows the architectures constructed by CGP-CNN (ConvSet) and CGP-CNN (ResSet). We can observe that these architectures are quite different; the summation and concatenation nodes are not used in CGP-CNN (ResSet), whereas these nodes are frequently used in CGP-CNN (ConvSet). These nodes lead the wide network; therefore, the network of CGP-CNN (ConvSet) is a wider structure than that of CGP-CNN (ResSet).

Added to this, we observe that CGP-CNN (ResSet) architecture has a similar feature with the ResNet \cite{he_deep_2016}. The ResNet consists of a repetition of two types of modules: the module with several convolutions with the shortcut connections without down-sampling, and down-sampling convolution with a stride of $2$.  Although our method cannot perform down-sampling in the ConvBlock and the ResBlock, we can see from Figure \ref{models} that CGP-CNN (ResSet) uses average pooling as an alternative to the down-sampling convolution. Furthermore, CGP-CNN (ResSet) has some convolutions with the shortcut connections, such as ResNet. Based on these observations, we can say that our method can also find the architecture similar to one designed by human experts, and that model shows a better performance.

Besides, while the ResNet has a very deep $110$-layer architecture, CGP-CNN (ResSet) has a relatively shallow and wide architecture.
We guess from this result that the number of output channels of ResBlock in the proposed method is one of the contributive parameters for improving the classification accuracy on the CIFAR-10 dataset.

For CGP-CNN (ResSet) on the default scenario, it takes about $14$ days to complete the optimization of the CNN architecture.
We observed a training time differs for each individual because various structures are generated by our method during the optimization.

\subsection{Result of the Small-data Scenario}
In the small-data scenario, we compare our method with VGG and ResNet.
We have trained VGG and ResNet models by the same setting of the re-training method in the proposed method; it is based on the training method of the ResNet \cite{he_deep_2016}.

\begin{figure}[t]
\includegraphics[width=0.9\linewidth]{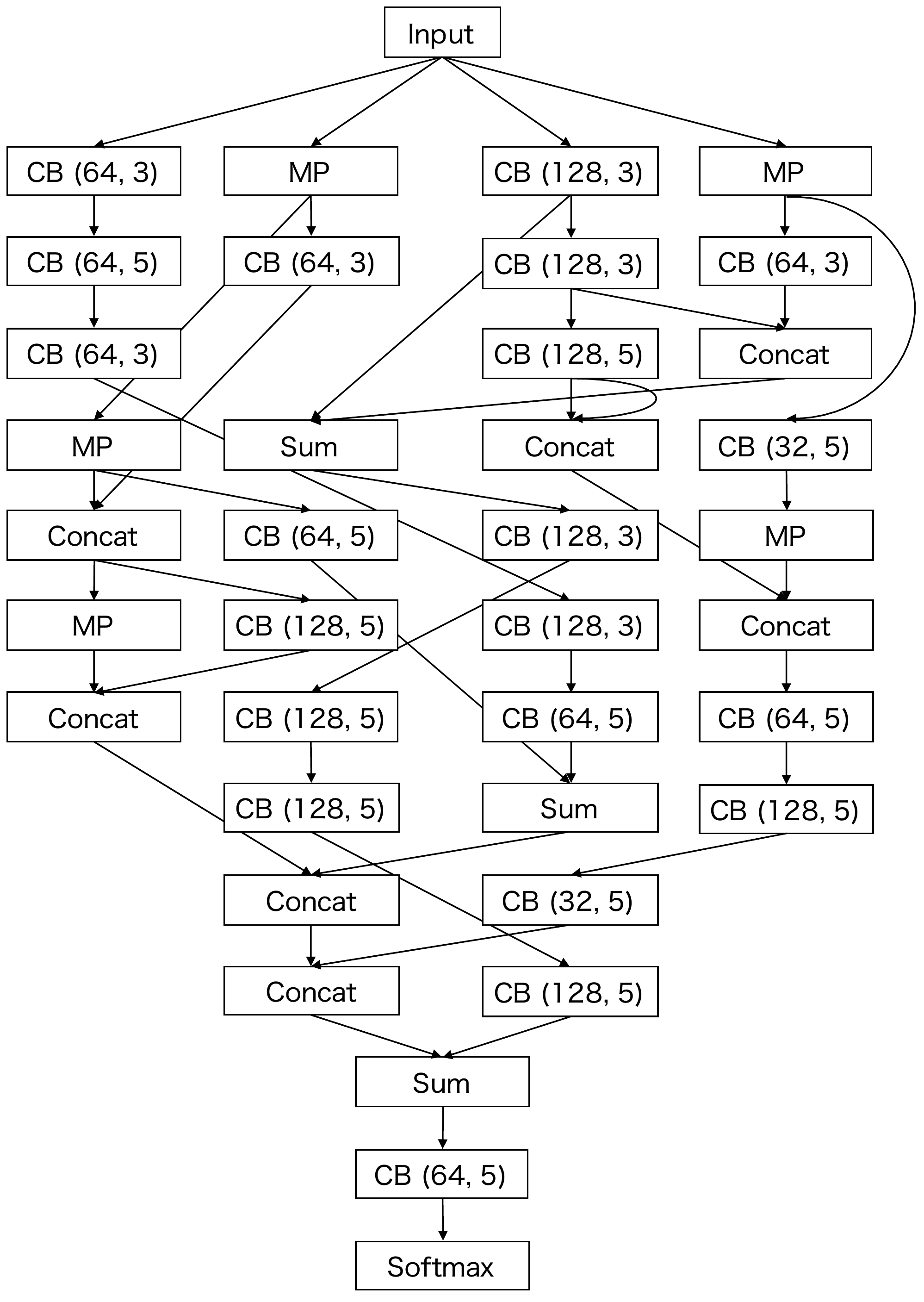}
\caption{The architecture of CGP-CNN (ConvSet) constructed in the small-data scenario.}
\label{model_small}
\end{figure}

Table 4 shows the comparison of error rates in the small-data scenario. We observe that our methods, CGP-CNN (ConvSet) and CGP-CNN (ResSet), can find better architectures than VGG and ResNet.
It is obvious that VGG and ResNet are inadequate for the small-data scenario because these architectures are designed for a relatively large amount of data. Meanwhile, our method can tune the architecture depending on the data size.
Figure \ref{model_small} illustrates the CGP-CNN (ConvSet) architecture constructed by using the proposed method.
As seen in Figure \ref{model_small}, our method has found a wider structure than that in the default scenario.

Additionally, we have re-trained this model with the $50,000$ training data and achieved a $8.05\%$ error rate on the test data. It suggests that the proposed method may be used to design a relatively good general architecture even with a small dataset.
For CGP-CNN (ResSet) on the small scenario, it takes about five days to complete the optimization of the CNN architecture.

\begin{table}[t]
  \caption{Comparison of error rates on the CIFAR-10 dataset (small-data scenario).}
  \label{results_small}
  \begin{tabular}{l|c|c} \hline
   Model & Error rate & \# params ($\times 10^6$) \\ \hline
   VGG \cite{simonyan_very_2014} & $24.11$ & 15.2 \\
   ResNet \cite{he_deep_2016} & $24.10$ & 1.7 \\ 
   CGP-CNN (ConvSet) & $23.48$  & $3.9$  \\
   CGP-CNN (ResSet) & $23.47$ & $0.83$ \\ \hline
  \end{tabular}
\end{table}

\section{Conclusion}
In this paper, we have attempted to take a GP-based approach for designing the CNN architectures and have verified its potential.
The proposed method constructs the CNN architectures based on CGP and adopts the highly functional modules, such as ConvBlock and ResBlock, for searching the adequate architectures efficiently.
We have constructed the CNN architecture for the image classification task with the CIFAR-10 dataset and considered two different data size settings. The experimental result showed that the proposed method could automatically find the competitive CNN architecture compared with the state-of-the-art models.

However, our proposed method requires much computational cost; the experiment on the default scenario needed about a few weeks in our machine resource. We can reduce the computational time if the training data are small (such as in the small-data scenario in the experiment). Thus, one direction of future work is to develop the evolutionary algorithm to reduce the computational cost of the architecture design, e.g., increasing the training data for the neural network as the generation progresses.
Moreover, to simplify the CNN architectures, we should consider to apply regularization techniques to the optimization process.
Also, it may be that we can manually simplify the obtained CNN architectures by removing redundant or less effective layers.
Another future work is to apply the proposed method to other image datasets and tasks.

\appendix

\section{Further Experiment in the Default Scenario}
In the GECCO 2017 paper, we reported the result of a single run for the proposed method. To produce a more reliable experimental result, we conducted the additional experiments. Here, we report the results over three independent runs with different random seed for each setting. Moreover, we compare the performance of obtained architectures by our method with ones by CoDeepNEAT \cite{miikkulainen2017evolving} and Large-scale Evolution \cite{real2017large} which are the architecture optimization methods published around the same time with our paper.

Table \ref{results_appendix} compares the performance of our method with the hand-crafted architectures and the state-of-the-art architecture optimization methods, summarizing the classification error rates, the numbers of learnable weight parameters in the obtained models, the computational time to complete the architecture optimization of the CNN, and the number of GPUs used in the experiment. For our method, the average values over three trials with different random seed are reported. The methods denoted by CoDeepNEAT \cite{miikkulainen2017evolving} and Large-scale Evolution \cite{real2017large} optimize the architectures using evolutionary computation. The error rate of the Large-scale Evolution shown in Table \ref{results_appendix} is the mean value over five trials reported in \cite{real2017large}.

As can be seen in Table \ref{results_appendix}, the error rates of our method are competitive with the state-of-the-art methods. In the evolutionary approaches, the Large-scale Evolution achieved the best error rate, but this approach used $250$ computers for the architecture optimization. Whereas, our method can find competitive architectures by using the reasonable machine resource. We guess that our method can reduce the search space and find the better architectures in early iterations by using the highly functional modules. We note that the comparisons among the architecture optimization methods are difficult because the experimental conditions and machine resources are different.

Table \ref{results_appendix2} shows the summary of the obtained architectures by our method in the six independent runs.
Across the six experiment runs, the best model has a testing error rate of $5.66\%$, and the worst model has a testing error rate of $6.81\%$.

\begin{figure*}[!h]
 \begin{minipage}[b]{0.45\linewidth}
  \centering
  \includegraphics[keepaspectratio, scale=0.32]{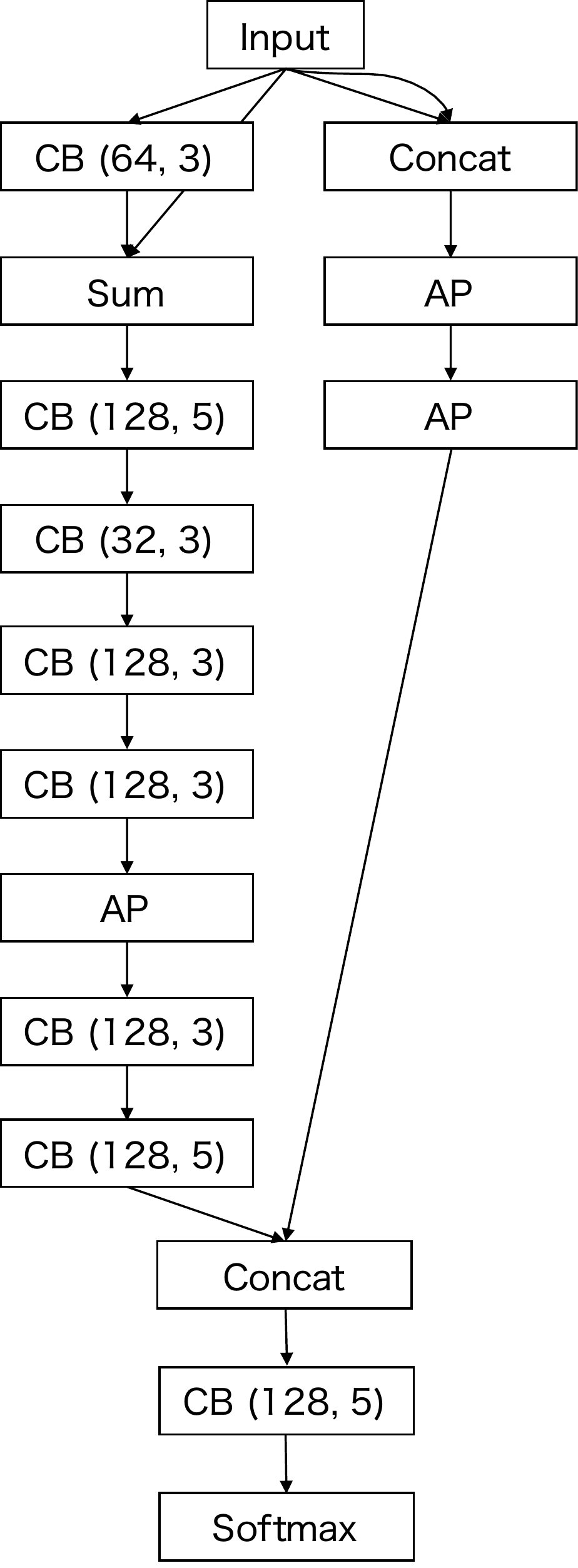}
  \subcaption{CGP-CNN (ConvSet)}\label{modelUA}
 \end{minipage}
 \begin{minipage}[b]{0.45\linewidth}
  \centering
  \includegraphics[keepaspectratio, scale=0.32]{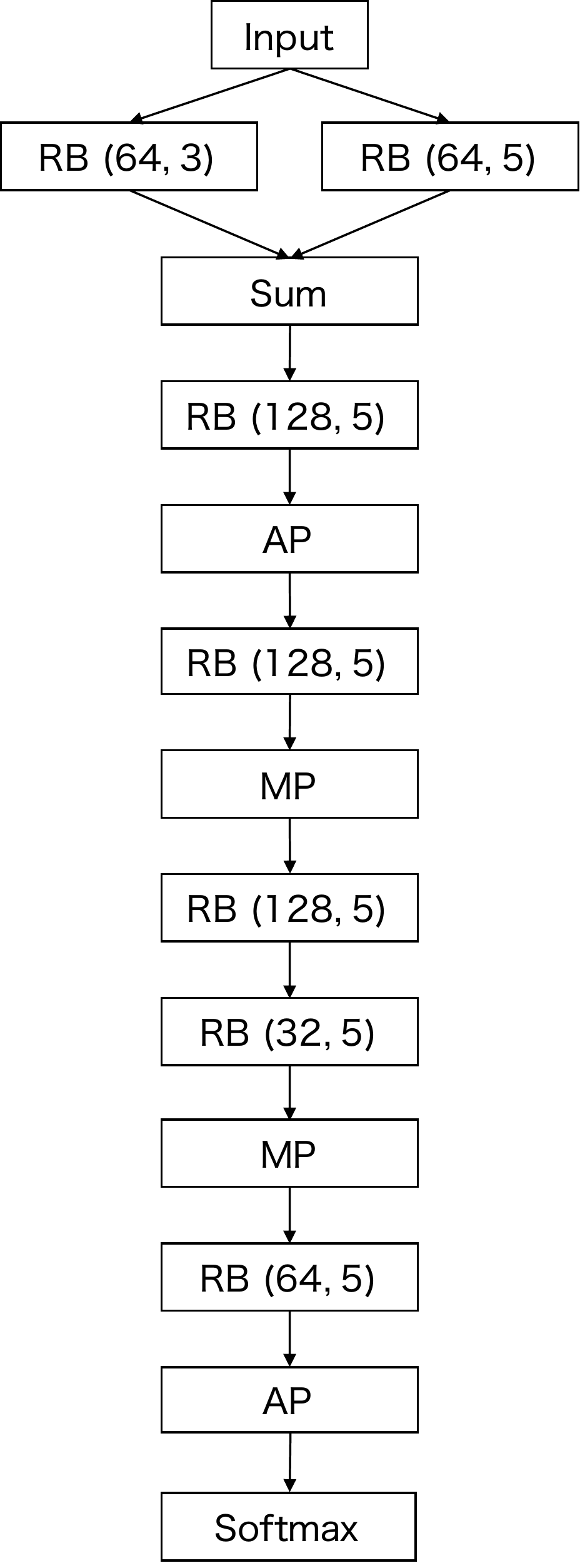}
  \subcaption{CGP-CNN (ResSet)}\label{modelUB}
 \end{minipage}
 \caption{The CNN architectures designed by our method on the default scenario. (a) CGP-CNN (ConvSet) achieved a $5.80\%$ error rate. (b) CGP-CNN (ResSet) achieved a $5.66\%$ error rate.}\label{modelsU}
\end{figure*}

Figure \ref{modelsU} shows the best architectures constructed by CGP-CNN (ConvSet) and CGP-CNN (ResSet).
We can observe that these architectures are relatively simple; the concatenation and summation nodes are not frequently used, and the networks are not wider structures.
In addition, the CGP-CNN (ResSet) architecture has a similar characteristic with the architecture shown in Figure \ref{modelB} and the ResNet \cite{he_deep_2016}; these architectures consist of a repetition of several convolutions and down-sampling.

\section{Detailed Experimental Settings and code}
The following experimental settings were missing in the GECOO 2017 paper:
\begin{itemize}
\item The maximum validation accuracy in the last 10 epochs is used as the fitness value for generated architecture.
\item The number of the active nodes in the individual of CGP is restricted between 10 to 50.  The mutation operation is re-applied until the restriction is satisfied.
\end{itemize}

The code of our method is available at https://github.com/sg-nm/cgp-cnn.

\begin{table*}[!h]
  \caption{Comparison of the error rates on the CIFAR-10 dataset (default scenario). The results of the hand- crafted architectures (Maxout, Network in Network, VGG, and ResNet), the state-of-the-art architecture optimization methods (MetaQNN, CoDeepNEAT, Large-scale Evolution, and Neural Architecture Search), and the proposed method (CGP-CNN) are displayed. The values are referred from the reference papers except for VGG and CGP-CNN. The average values over three trials with different random seed are reported for CGP-CNN.}
  \label{results_appendix}
  \begin{tabular}{l|c|c|c|c} \hline
   Model & Error rate & \# params ($\times 10^6$) & Optimization time & \# GPUs \\ \hline
   Maxout \cite{goodfellow_maxout_2013} & $9.38$ & -- & -- & -- \\ 
   Network in Network \cite{lin_network_2014} & $8.81$ & -- & -- & -- \\ 
   VGG \cite{simonyan_very_2014} & $7.94$ & $15.2$ & -- & -- \\ 
   ResNet \cite{he_deep_2016} & $6.61$ & $1.7$ & -- & -- \\ 
   MetaQNN \cite{baker_designing_2016} & $9.09$ & $3.7$ & 8--10 days & 10 \\
   CoDeepNEAT \cite{miikkulainen2017evolving}  & $7.30$ & -- & -- & -- \\
   Large-scale Evolution \cite{real2017large}  & $5.9$ & -- & about 10 days & 250 computers \\
   Neural Architecture Search \cite{zoph_neural_2016} & $3.65$ & $37.4$ & -- & 800 \\
   CGP-CNN (ConvSet) & $6.34$ & $ 1.75 $ & $15.2$ days & 2 \\
   CGP-CNN (ResSet) & $6.05$ & $ 2.64 $ & $13.7$ days & 2 \\ \hline
  \end{tabular}
\end{table*}

\begin{table*}[!h]
  \caption{The summary of the obtained architectures by our method in the six independent runs. The result recording the error rate of 5.66 was run on two NVIDIA TITAN X GPUs. Other experiments were run on two NVIDIA GeForce GTX 1080 GPUs.}
  \label{results_appendix2}
  \begin{tabular}{l|c|c|c} \hline
   Model & Error rate & \# params ($\times 10^6$) & Optimization time  \\ \hline
   CGP-CNN (ConvSet) & $6.81$ & $1.81$ & $13.8$ days  \\ 
   CGP-CNN (ResSet) & $6.51$ & $3.41$ & $15.8$ days  \\ 
   CGP-CNN (ConvSet) & $6.41$ & $1.95$ & $19.4$ days  \\ 
   CGP-CNN (ResSet) & $5.98$ & $1.68$ & $14.9$ days  \\ 
   CGP-CNN (ConvSet) & $5.80$ & $1.50$ & $12.0$ days  \\
   CGP-CNN (ResSet)  & $5.66$ & $2.84$ & $10.4$ days  \\ \hline
  \end{tabular}
\end{table*}

\bibliographystyle{ACM-Reference-Format}
\bibliography{sample-sigconf}

\end{document}